\documentclass[twocolumn,twocolumn]{IEEEtran}
\usepackage[T1]{fontenc}
\usepackage{color}
\usepackage{float}
\usepackage{mathrsfs}
\usepackage{amsmath}
\usepackage{amssymb}
\usepackage{graphicx}
\usepackage[unicode=true,
 bookmarks=true,bookmarksnumbered=true,bookmarksopen=true,bookmarksopenlevel=1,
 breaklinks=false,pdfborder={0 0 0},pdfborderstyle={},backref=false,colorlinks=false]
 {hyperref}
\hypersetup{pdftitle={Your Title},
 pdfauthor={Your Name},
 pdfpagelayout=OneColumn, pdfnewwindow=true, pdfstartview=XYZ, plainpages=false}

\makeatletter

\floatstyle{ruled}
\newfloat{algorithm}{tbp}{loa}
\providecommand{\algorithmname}{Algorithm}
\floatname{algorithm}{\protect\algorithmname}

\usepackage[caption=false,font=footnotesize]{subfig}
\usepackage{algorithm}
\usepackage{algorithmic}

\usepackage{multirow} 
\usepackage{amsmath} 
\usepackage{xcolor}

\allowdisplaybreaks[4]

\ifCLASSOPTIONcompsoc
\usepackage[caption=false,font=normalsize,labelfont=sf,textfont=sf]{subfig}
\else
\usepackage[caption=false,font=footnotesize]{subfig}
\fi

\usepackage{tabularray}
\usepackage{cite}
\usepackage{bm}
\usepackage{algorithmic}
\usepackage{algorithm}
\usepackage{graphicx}
\IEEEoverridecommandlockouts

\usepackage{lettrine}

\usepackage{geometry}
\@ifundefined{showcaptionsetup}{}{%
 \PassOptionsToPackage{caption=false}{subfig}}
\usepackage{subfig}
\makeatother

\begin{document}

\title{\textcolor{black}{Robust UAV Path Planning with 
Obstacle Avoidance for Emergency Rescue}}

\author{\IEEEauthorblockN{Junteng Mao$^{\dagger}$, Ziye Jia$^{\dagger}$$^{\ast}$, Hanzhi Gu$^{\dagger}$,
Chenyu Shi$^{\dagger}$, Haomin Shi$^{\dagger}$, Lijun He$^{\star }$ and Qihui Wu$^{\dagger}$\\}%
\IEEEauthorblockA{$^{\dagger}$The Key Laboratory of Dynamic Cognitive 
System of Electromagnetic Spectrum Space, Ministry of Industry and Information 
Technology, Nanjing University of Aeronautics and Astronautics, Nanjing, Jiangsu, 
211106, China\\
$^{\ast}$National Mobile Communications Research Laboratory, Southeast University, 
Nanjing, Jiangsu, 211111, China\\
$^{\star }$The School of Information and Control Engineering, China University of Mining and Technology, 
Xuzhou 221116, China\\
\{twistfate, jiaziye, hanzhi.gu, chenyu\_Shi, shihaomin, wuqihui\}@nuaa.edu.cn, lijunhe.xd@gmail.com}
\thanks{This work was supported in part by National Natural Science 
Foundation of China under Grant 62301251 and 62201463, in part by the 
Natural Science Foundation on Frontier Leading Technology Basic 
Research Project of Jiangsu under Grant BK20222001, in part by  
the open research fund of National Mobile 
Communications Research Laboratory, Southeast University (No. 2024D04), 
in part by the Aeronautical Science Foundation of China 2023Z071052007, 
and in part by the Young Elite Scientists Sponsorship Program by CAST 
2023QNRC001. (\textit{Corresponding author: Ziye Jia.})}}

\maketitle
\thispagestyle{empty}

\pagestyle{empty} 
\begin{abstract}
The unmanned aerial vehicles (UAVs) are efficient tools for 
diverse tasks such as electronic reconnaissance, agricultural 
operations and disaster relief. In the complex three-dimensional (3D) 
environments, the path planning with obstacle avoidance for UAVs is a 
significant issue for security assurance. In this paper, we construct a 
comprehensive 3D scenario with obstacles and no-fly zones for dynamic UAV 
trajectory. Moreover, a novel artificial potential field algorithm coupled with 
simulated annealing (APF-SA) is proposed to tackle the robust path 
planning problem. APF-SA modifies the attractive and repulsive potential 
functions and leverages simulated annealing to escape local minimum and converge 
to globally optimal solutions. Simulation results 
demonstrate that the effectiveness of APF-SA, enabling efficient autonomous path planning for
UAVs with obstacle avoidance. \par
\end{abstract}

\newcommand{\CLASSINPUTtoptextmargin}{0.8in}

\newcommand{\CLASSINPUTbottomtextmargin}{1in}

\section{Introduction}

\lettrine[lines=2]{I} {n} recent \textcolor{black}{years, unmanned 
aerial vehicles (UAVs) play increasingly 
important roles in applications of military mission, natural 
exploration, disaster relief, etc\cite{10559211}.}\textcolor{black}{{} However, 
when the UAV flies autonomously in complex and changeable environments, 
how to effectively avoid obstacles and ensure flight safety becomes an 
urgent problem. Besides, in the process of searching unknown and complex 
environments, the 
finite energy capacity of UAVs imposes constraints on their service 
capabilities. To optimize mission execution, the autonomous path planning 
mechanisms are essential. Meanwhile, in order to accomplish multifarious 
monitoring, mapping, rescuing and searching tasks efficiently and safely, 
it is necessary for the UAV to have comprehensive obstacle avoidance 
abilities\cite{DONG2024}\cite{10599389}. Therefore, it is significant to study the 
autonomous path planning algorithm for UAVs with obstacle avoidance 
abilities.}

\textcolor{black}{Most existed researches on autonomous path planning 
and obstacle avoidance for UAVs are based on known environments. 
Traditional methods are classified as classic and heuristic approaches 
such as vector field histogram method and various geometry search 
algorithms (e.g. rapidly-exploring random tree star algorithm \cite{9970002}, 
Dijkstra's algorithm, Floyd's algorithm, A* algorithm, ant colony 
algorithm (ACO)\cite{10225633}, etc.). For example, 
authors in \cite{6561444} 
proposed an improved ACO algorithm using a three-dimensional (3D) grid 
and a new weight-climbing parameter, which solved the problems of 
premature convergence and low efficiency of the traditional ACO 
algorithm. In \cite{10295735}, the authors used an improved A* algorithm to 
shorten the search cost and reduce the amount of search and inflection 
points successfully, making the path smoother and safer. Nevertheless, 
most algorithms suffer from overlong runtimes, excessive iterations, 
and low real-time processing capability. Therefore, it is essential to 
design a better algorithm to solve the obstacle avoidance and path 
planning problems for these complex environments.}

\textcolor{black}{The traditional artificial potential field (APF) 
method is widely used in UAV trajectory planning due to its advantages 
of simple calculations, outstanding real-time control, smooth path, 
laconic implementation, and convenient debugging\cite{9830995}. However, 
UAVs are apt to be trapped into a local minimum and 
oscillations problem with traditional APF\cite{10250980}. Besides, most 
researches on APF focused on two-dimensional environments, ignoring the 
shape and distribution of the obstacles in 3D terrain environments, 
especially when dealing with dynamic obstacles. In this 
regard, authors in \cite{10033554} presented a UAV trajectory planning 
method utilizing a goal-biased APF in combination with the Rapidly 
exploring Random Tree star (RRT*) algorithm to improve the convergence 
speed and minimize the number of iterations. In \cite{9538804}, authors
presented a rotating potential field methodology that enabled UAVs to 
avoid common local minima and oscillations effectively. However, these 
works fail to achieve smooth path planning and ignore the 3D dynamic 
environment.}

\textcolor{black}{Hence, in this work, we consider an unknown, complex 
forest firefighting environment with both dynamic and static 
obstacles. In particular, we model the ground obstacles, overhead 
obstacles, and dynamic obstacles that UAVs may encounter during its 
flight in the forest as cylinders, static spheres, and moving spheres, 
respectively. Then, we propose a novel artificial potential field 
algorithm based on simulated annealing (APF-SA), which significantly 
improves 
the conventional APF. Specifically, it amplifies the gravity 
function around the target point and introduces a repulsion correction 
factor to optimize the path towards the firefighting destination. 
Furthermore, the integration of simulated annealing (SA) techniques 
addresses the local minimum issue, enabling the algorithm to converge 
towards global optimal solutions. APF-SA aims to achieve 
fast firefighting safely and effectively through 
path planning autonomously with obstacle avoidance. Finally, 
we conduct simulations to verify the effectiveness of the proposed 
method.}

The rest of the paper is organized as follows. The system model and 
problem formulation are presented in Section \ref{sec:System-Model}. 
Then, the APF-SA algorithm
is proposed in Section \ref{sec:Algorithm-Design}. Simulations 
are conducted in Section \ref{sec:Simulation-Results}, and finally
conclusions are drawn in Section \ref{sec:Conclusions}.

\section{System Model and Problem Formulation\label{sec:System-Model}}

\subsection{3D Forest Scenario}
\begin{figure}[tb]
\centering

\includegraphics[scale=0.33]{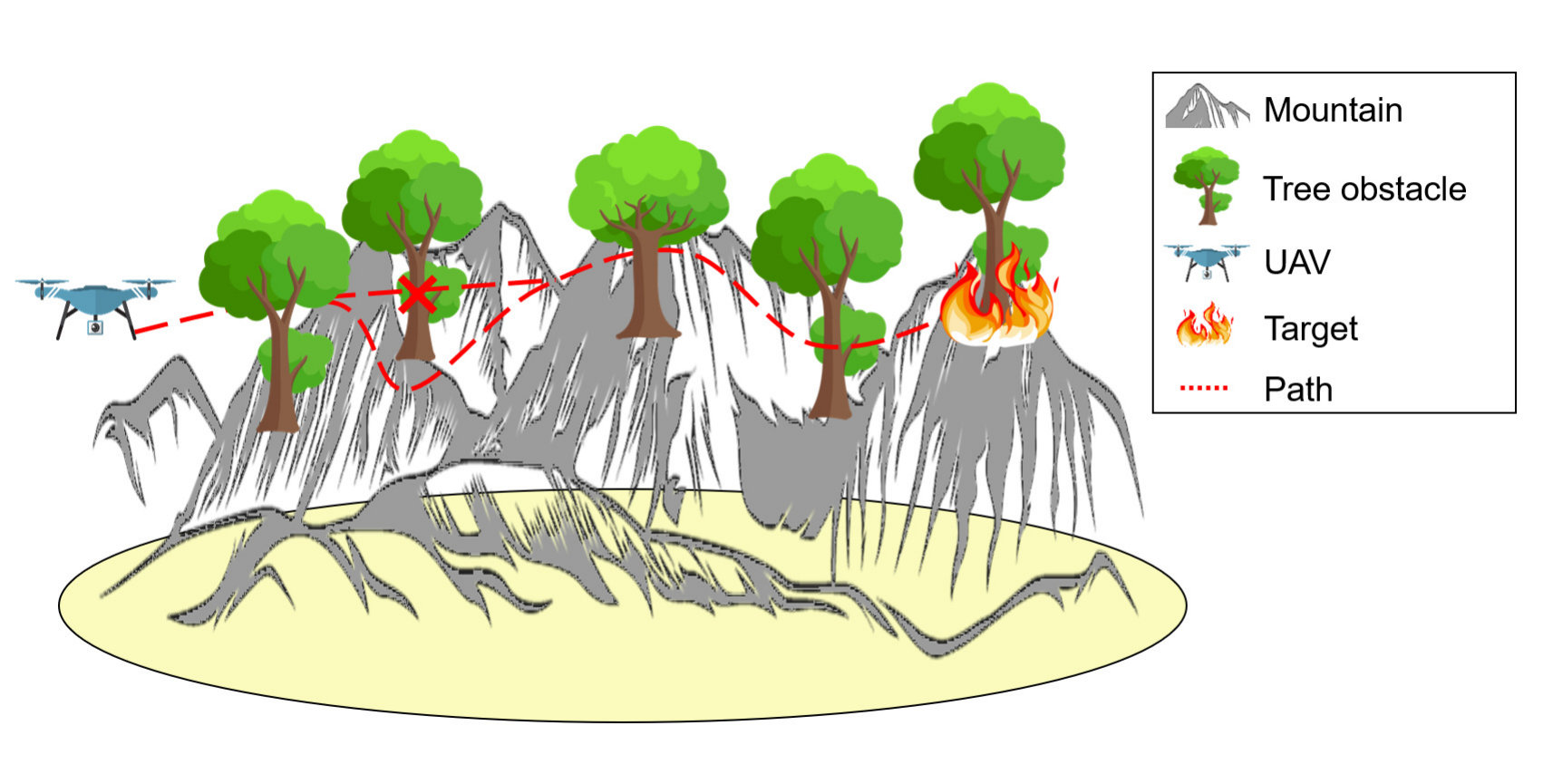}\vspace{-2mm}\textcolor{black}{\caption{\label{fig:scence}Obstacle avoidance trajectory scenario for UAV in 3D environment.}}

\vspace{-5mm}
\end{figure}

\begin{figure}[t]
\centering

\includegraphics[scale=0.39]{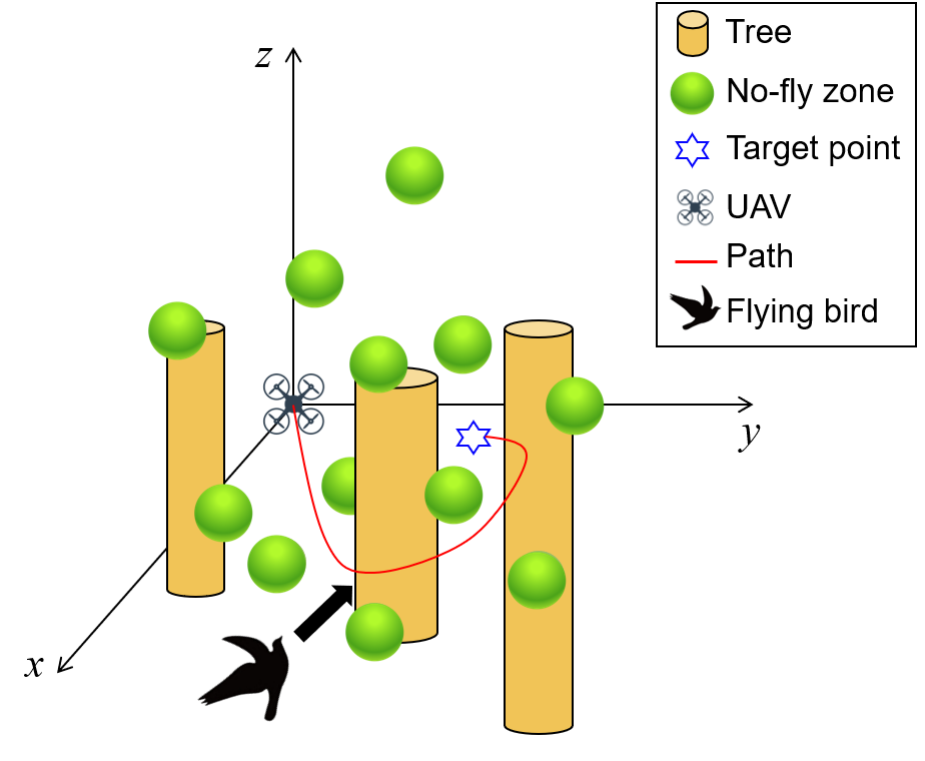}\vspace{-2mm}\textcolor{black}{\caption{\label{fig:system model}Path planning model for UAV.}}

\vspace{-5mm}
\end{figure}

As shown in Fig. \ref{fig:scence}, the UAV needs to perform 
the task of detecting fire in the mountains. We denote the 3D coordinate 
of target point, initiation point and UAV position as $X_g(x,y,z)$, $X_q(x,y,z)$ and $X_u(x,y,z)$, respectively. 
Considering the complex conditions in 
mountainous areas during the flight, the UAV may encounter a series of 
different obstacles $\textbf{\textit{O}}=(X_{o1},X_{o2},\dots,X_{oi},\dots,X_{ol})$, 
including trees, no-fly zones, and other
objects, where $X_{oi}(x,y,z)$ represents the barycenter of the $i$-th obstacle, and $l$ denotes 
the total number of obstacles. In order to characterize the obstacles and 
perform 3D obstacle avoidance, different obstacles should be modeled independently. 
As for the obstacles of trees, we conceive them as cylinders with base 
radius $r_1$ and height $h$. 
The no-fly zone and other flying objects, are represented by spheres with 
radius $r_2$ and $r_3$, respectively. Moreover, the positions of the 
flying objects' centers change over time, as shown in Fig. \ref{fig:system model}. 
In addition, the 
speed of the UAV during the flight changes with the variation of the resultant force. 
However, there exist upper limitation to the power of UAV, so 
the speed range is $[v_{min},v_{max}]$. In order to simplify 
the motion model of the UAV, the basic motion unit is set as a fixed step size $\Delta s$. 
If the driving direction of the UAV is $\delta (\gamma_x, \gamma_y, \gamma_z)$, the 
position update formula of the UAV is \vspace{-2mm}

\begin{equation}
\left\{
\begin{aligned}
x'=x+{\Delta  s}*{\gamma_x},\\
y'=y+{\Delta  s}*{\gamma_y},\\
z'=z+{\Delta  s}*{\gamma_z},
\end{aligned}
\right.
\end{equation}
wherein, $\gamma_x$ represents the angle between 
the projection on the ground of the UAV's flight direction and the X-axis, 
while $\gamma_y$ and $\gamma_z$ indicate the angle with the Y-axis and the Z-axis, respectively.

Therefore, a series of track points can be obtained according to the preset 
step size and moving direction of the UAV.
Moreover, we adopt the grid 
method to describe the 3D scenario, denoted as $\mathcal{P} $, which is 
a cuboid region of size $l_x\times l_y\times l_z$, consisting of several small cuboids with side 
length $\Delta l$. Hence, the number of small cubes is \vspace{-2mm}

\begin{equation}
N_x=\frac{l_x}{\Delta l},\;N_y=\frac{l_y}{\Delta l},\;\text{and} \;N_z=\frac{l_z}{\Delta l}.
\end{equation}
It is obvious that the hyperparameter $\Delta l$ plays an important role 
in the whole grid construction, since the decrement of $\Delta l$ results in an 
increment of $N_x$, $N_y$ and $N_z$, and the 
whole 3D scenario can be more refined \cite{10404342}.
However, once the 
whole scenario reaches a certain scale, slightly reducing $\Delta l$ 
can cause significant increment in the calculation cost, which is 
unacceptable. Therefore, it is significant to balance the size of the scenario 
with $\Delta l$ to achieve detailed scene construction 
and acceptable calculation amplitude.
Moreover, the position of the UAV should always be perceived for collision detection.
In detail, $\rho(X_u,X_{oi}) $ represents the distance between the UAV $X_u$ and obstacle $X_{oi}$, 
i.e., \vspace{-2mm}
\begin{equation}
\rho(X_u,X_{oi})=\sqrt{(X_u-X_{oi})^2}.
\end{equation}
Further, flag $C_i$ represents collisions, i.e,\vspace{-2mm}

\begin{equation}
C_i=
\begin{cases}
    1,&\rho(X_u,X_{oi})\leq r_i,\\
    0,&\rho(X_u,X_{oi})\textgreater r_i,
\end{cases}
\end{equation}
where $r_i$ is the radius of obstacles. $C_i$ = 1 indicates there is 
a collision between the UAV and an obstacle, and $C_i$ = 0 otherwise. 

\subsection{Energy Cost Model}
The finite maximum energy capacity $W_{max}$ of the UAV during a 
mission imposes constraints on its payload and service capabilities \cite{9184929}.
Therefore, to ensure that the UAV can return smoothly during the execution 
of the task, we have: \vspace{-2mm}

\begin{equation}
\sum_{X_u=X_q}^{X_n}{\Delta s}\cdot {\Delta w}\leq W_{max},
\end{equation}
where $\Delta w$ represents the energy consumed per unit length of the UAV 
trajectory and $X_n$ is the coordinate position of the last point in the flight path 
of the UAV.

\subsection{Problem Formulation}
Since the purpose of the forest fire fighting problem is to reach the target 
as soon as possible by rationally planning the path of the UAV, the goal is 
to minimize the length of the route. Therefore, the optimization 
problem is formulated as:
\begin{alignat}{2}
    \mathop{\min}_{X_u}\quad& \sum_{X_u=X_q}^{X_n}{\Delta s}&\\
    \mbox{s.t.}\quad
    &S \in \mathcal{P}, \\
    &\sum_{X_u=X_q}^{X_n} C_i = 0.
\end{alignat}
$X_n$, $X_q$ and $X_u$ represent the final location, start point, 
and the latest location of the UAV respectively.
Therefore, we need to design a series of points that meet the 
requirements and obtain a short trajectory.

\section{Algorithm Design\label{sec:Algorithm-Design}}
\subsection{Basic of APF\label{sec:Basic of APF}}

APF was first proposed by Khatib in 1986,
and its basic principle is to simulate the motion of particles in a 
virtual potential field\cite{10169744}. In this mechanism, the UAV is regarded 
as a 
particle, and then a gravitational field is set up around the target 
point from the UAV to the target. Generally, the gravity becomes stronger
as the distance increases, attracting the UAV to move towards the 
target point. Meanwhile, a suitable repulsive force field is 
established around obstacles. The direction of repulsive force is from obstacles to
the UAV, and its strength is usually inversely proportional to the 
distance of the UAV from the obstacle, 
allowing the UAV to avoid collisions with obstacles or no-fly zones. 
\begin{figure}[tb]
\centering

\includegraphics[scale=0.37]{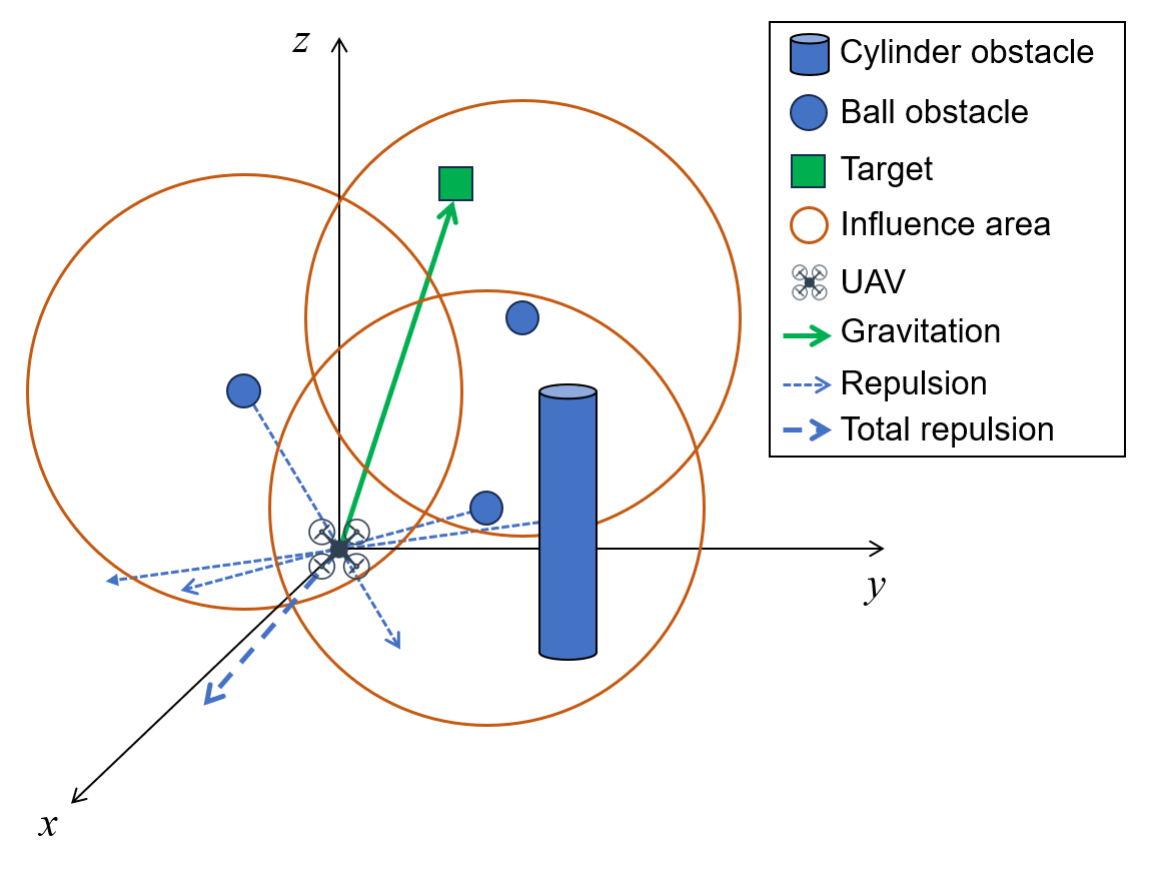}\vspace{-2mm}
\begin{flushright}
\textcolor{black}{\caption{\label{fig:principle}APF principle.}   } 
\end{flushright}

\vspace{-5mm}
\end{figure}
As shown in Fig. \ref{fig:principle}, the potential field function 
of APF 
can be expressed as the sum of the gravitational potential field 
and the repulsion potential field, i.e.,\vspace{-2mm}

\begin{equation}
U(X_u)=U_{att}\ (X_u)+U_{rep}\ (X_u),
\end{equation}
where $U_{att}\ (X_u)$ is the gravitational potential field, and $U_{rep}\ (X_u)$ 
is the repulsive potential field. The corresponding gravitational or 
repulsive function can be obtained by finding the negative gradient 
of the potential field function. Since the motion of the UAV 
is constrained by the gravitational and repulsive forces, 
the resultant force can be expressed as:\vspace{-2mm}

\begin{equation}
F_s(X_u)=F_{att}\ (X_u)+F_{rep}\ (X_u),
\end{equation}
where $F_{att}\ (X_u)$ denotes the gravitational force, 
directed by the UAV towards the target point, 
so that the UAV can approach the target. $F_{rep}\ (X_u)$ represents the 
repulsive force, and the direction is the reverse direction of the 
connection between the UAV and obstacles, enabling the UAV to avoid obstacles.

In order to further analyze the force on the UAV, the repulsive force 
and the attractive force can be calculated separately. The gravitational 
potential field is expressed as $U_{att}\ (X_u)$, whose size is 
determined by the relative distance between the UAV and the target point, 
and it always pulls the UAV to move towards the target point, i.e., \vspace{-2mm}

\begin{equation}
U_{att}(X_u) =\frac{1}{2} \eta  (X_u-X_g )^2,
\end{equation}
in which $\eta$ represents the attractive force gain coefficient. On this basis, the 
negative gradient of the gravitational potential field function can be 
obtained as:\vspace{-2mm}

\begin{equation}
F_{att}(X_u)=-\eta  (X_u-X_g ).
\end{equation}
Similarly, the repulsive force potential field of obstacles to the UAV is 
$U_{rep}\ (X_u)$. The repulsive force increases when the UAV gets close to the 
obstacles. As such, the obstacle avoidance can be realized, and the repulsive 
field function is \vspace{-2mm}

\begin{equation}
    U_{rep}\ (X_u)=
    \begin{cases}
        \frac{1}{2} \beta (\frac{1}{\rho(X_u,X_{oi})}-\frac{1}{\rho _{0}})^2,&\text{$\rho(X_u,X_{oi})<\rho _0$},\\
        0,&\text{$\rho(X_u,X_{oi})\geq\rho _0$,}
    \end{cases}
\end{equation}
in which $\beta$ is the repulsion gain constant, and $\rho _0$ indicates 
the maximum impact distance of a single obstacle.

\subsection{Modified Potential Field Function\label{sec:Modified Potential Field Function}}
The main contributing factor for the unreachable target is that when approaching 
the target point, the gravitational force exerting on the UAV is too 
small, and the repulsive force received is too large, causing the UAV 
to yaw. Therefore, based on the improved methods in \cite{9832753}, \cite{9734752}, 
APF-SA is decided to modify the gravitational function and repulsion 
function. Under the action of the original gravity function, although the 
UAV is close to the target point, the gravity also reduced. 
Therefore, when there are obstacles near the target point, the repulsive 
force received by the UAV may be greater than the gravitational force, 
resulting in the unreachable target point. Hence, it is necessary 
to increase the gravity appropriately near the target point without 
affecting other regions, and the corrected attractive function is \vspace{-2mm}

\begin{equation}
F_{att}(X_u)=-\eta  [\rho (X_u,X_g )+e^{-(\rho(X_u,X_g)-\varepsilon \times \Delta s)}],
\end{equation} 
wherein, $\eta$ is the attractive force gain, $\varepsilon$ is the step size impact factor 
and $\rho (X_u,X_g )$ represents 
the relative 
distance between UAV $X_u$ and goal point $X_g$. 

\begin{figure}[t]
\centering

\includegraphics[scale=0.5]{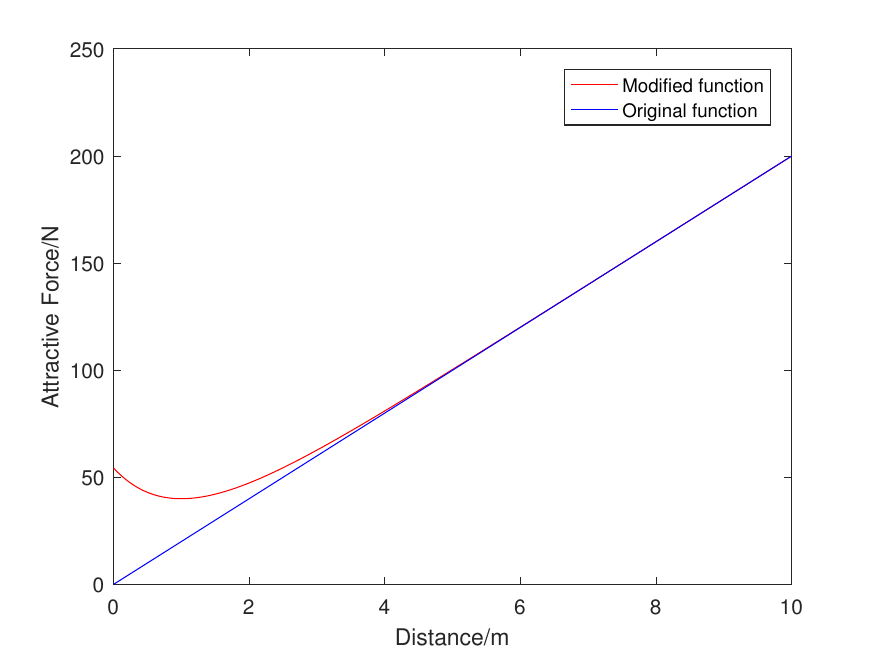}\vspace{-2mm}\textcolor{black}
{\caption{\label{fig:attractive force}Comparison of attractive force function original v.s. modified.}
}

\vspace{-3mm}
\end{figure}

As shown in Fig. \ref{fig:attractive force}, the attractive force intensifies
as the UAV gets close to the target, enabling a successful arrival.
When the UAV is located far from the target, its gravity
remains largely consistent with the original gravity function. 
These observations validate the effectiveness of the revised gravity function. 
Meanwhile, the repulsion function is optimized 
and $\rho (X_u,X_g )$ is introduced, so that the repulsion force is not 
only related to the distance between the UAV and the obstacle, but also 
related to the distance between the UAV and target point. When the UAV gets close
to the target point after the correction, the repulsive force of the 
obstacle gradually approaches zero. Hence, the target point becomes 
the lowest potential energy point, and the revised 
repulsion function is: \vspace{-2mm}

\begin{equation}
    U_{rep}(X_u)\!\!=\!\!
    \begin{cases}
        \!\frac{1}{2} \beta (\frac{1}{\rho {(X_u,X_{oi})}}\!\!-\!\!\frac{1}{\rho _{0}})^2 {\rho (X_u,X_g)}^\mu,&\!\!\!\!\!\!\text{$\rho {(X_u,X_{oi})}\!\!<\!\!\rho _0$},\\
        \!0,&\!\!\!\!\!\!\!\text{$\rho {(X_u,X_{oi})}\!\!\geq\!\!\rho _0$},
    \end{cases}
\end{equation} 
where $U_{rep}\ (X_u)$ is the revised repulsion field function. $\rho _0$ indicates the repulsion influence 
radius of the obstacle, and $\mu $ represents a positive real number, 
indicating the repulsive force near the target point 
is small enough, while the attractive force can still pull the UAV, 
thus solving the unreachable problem of the target point. As the 
negative gradient of the improved repulsion function, the repulsive force 
function is \vspace{-2mm}

\begin{equation}
    F_{repg}\ (X_u)=
    \begin{cases}
        F_{repg1}+F_{repg2},&\text{$\rho {(X_u,X_{oi})}\leq\rho _0$},\\
        0,&\text{$\rho {(X_u,X_{oi})}> \rho _0$},
    \end{cases}
\end{equation} 
wherein, \vspace{-2mm}

\begin{equation}
    F_{repg1}= \beta (\frac{1}{\rho {(X_u,X_{oi})}}- \frac{1}{\rho _0})\frac{{\rho (X_u,X_g)}^\mu}{{\rho {(X_u,X_{oi})}}^2}\frac{\partial\rho {(X_u,X_{oi})}}{\partial x},
\end{equation}
and\\ 
\begin{equation}
        F_{repg2}\!=\! \frac{\mu }{2}\beta (\frac{1}{\rho {(X_u,X_{oi})}}- \frac{1}{\rho _0})^2 \frac{\partial\rho {(X_u,X_g)}}{\partial x}{\rho (X_u,X_g)}^{\mu-1}.
\end{equation} 
The original repulsive force $F_{repg1}$ points to the UAV from 
the obstacle. In order to avoid excessive repulsion, which causes the 
resultant force to deviate from the target point, a new repulsion force $F_{repg2}$ is 
added, which points toward the target point.

\subsection{\textcolor{black}{APF-SA Design}}
As mentioned in section \ref{sec:Basic of APF}, the UAV is apt to fall 
into a local minimum area in the process of path planning. During the 
flight, there is a point where the gravitational force and the 
repulsive force are equal in magnitude but opposite in direction. This 
results in a zero resultant force on the UAV, causing it to lose 
directional information for its next movement. Consequently, the 
navigation algorithm fails, preventing the UAV from reaching its 
target. For this reason, the simulated annealing (SA) algorithm is 
adopted to randomly generate a new reachable trajectory point near the 
current position of the UAV, and make the UAV fly to this trajectory 
point. In this circumstance, the original force balance is broken, so 
that the UAV can break the local minimum dilemma and continue to move 
towards the target point.
\par SA  is a generic probabilistic metaheuristics, 
initially introduced by Kirkpatrick $\textit{et al.}$\cite{9371396}. 
The algorithm is divided into an external cycle and an 
internal cycle, where the external cycle represents a drop in temperature, 
and the internal cycle is a different state at this temperature due to multiple 
random disturbances. The inner loop accepts the new state according to the 
Metropolis criterion \cite{10079729}, and the probability that the particle tends to equilibrium 
at temperature $T$ is $p=e^{(-\frac{\Delta E}{kT})}$, where $E$ is the internal 
energy at temperature $T$, $\Delta E$ denotes the internal energy of change, 
and $k$ represents the Boltzmann constant \cite{9534874}. In detail, the Metropolis criterion is\vspace{-2mm}

\begin{equation}
    p=
    \begin{cases}
        e^{-\frac{E(t_n)-E(t_0)}{T}},&\text{$E(t_n) > E(t_0)$,}\\
        1,&\text{$E(t_n) \leq E(t_0)$,}
    \end{cases}
\end{equation} 
where $t_n$ is the last time slot, and $t_0$ indicates the previous time slot. 
$E(t_0)$ is the internal energy at $t_0$, $E(t_n)$ is the internal energy 
at $t_n$, and $T$ stands for the current temperature.
In the SA algorithm, the temperature $T$ decreases as,

\begin{equation}
    T(t) = \alpha T(t-1),
\end{equation} 
wherein, $\alpha$  is a positive real number slightly less than 1, whose value is 
usually within (0.85, 1), and $t$ is the number of iterations.

\par The specific process APF-SA is shown in Algorithm \ref{alg:APF-SA}. 
The target point provides the attractive force for the UAV to fly close to 
the target, while the obstacles along the way provide a certain repulsive 
force to help the UAV successfully avoid obstacles (line \ref{alg:Initialization}). When the UAV is near 
the obstacle, it determines whether it enters the local minimum value (line \ref{alg:judgement}). 
If it is not trapped, it flies through smoothly (lines \ref{alg:calculate}-\ref{alg:new point}). If it is trapped, it 
jumps into the simulated annealing process (lines \ref{alg:equal}-\ref{alg:SA}). 
A new locus point $X^{'}$ is randomly generated at the current local 
minimum point $X_u$ (line \ref{alg:add}), and then it is judged whether the random point is located 
in the obstacle region. If not, the potential field at points $X_u$ and 
$X^{'}$ is calculated respectively according to the potential field function (line \ref{alg:test}). 
If the potential field of the random point is lower than that of the local 
minimum point, point $X^{'}$ is accepted as the next point (line \ref{alg:test}). Otherwise, the 
point is accepted as the next point with probability $p$. This process
keeps running until the end of the flight (line \ref{alg:final}).
\vspace{-3mm}
\begin{algorithm}[t]
\caption{APF-SA.\label{alg:APF-SA}}

\begin{algorithmic}[1]

\REQUIRE Starting point and target point.

\ENSURE Planned path and time consumption $t$.

\STATE\textit{Initialization:\label{alg:Initialization}} ${t}=0$, start point $ X_q$, target point $ X_g$, UAV position $X_u$, target point produces ${F_{att}}$, and obstacles produce ${F_{rep}}$.

\REPEAT

\STATE\label{alg:judgement} Calculate resultant force: ${F_{s}}$ = ${F_{att}}$ + ${F_{rep}}$.

\IF{${F_{s}}$ = ${0}$} \label{alg:equal}

\STATE\label{alg:SA} Jump into SA (lines \ref{alg:add}-\ref{alg:test}).

\REPEAT

\STATE\label{alg:add} Add a random action point.

\STATE\label{alg:test} Test whether the new point meets the requirements.

\UNTIL the new point is successful.

\STATE\label{alg:success} Target the new point to escape local minima.

\ELSE

\STATE\label{alg:calculate} Calculate total force.

\STATE\label{alg:new point} Obtain new action point by step size.

\ENDIF 

\STATE Move to the next point.

\STATE Update ${X_u}$.

\UNTIL\label{alg:final}${X_u}={X_g}$.

\end{algorithmic}
\end{algorithm}

\section{Simulation Results\label{sec:Simulation-Results}}

Simulations are conducted in the following scenario: the UAV is treated 
as a particle and a $\textrm{200m}\times\textrm{200m}\times\textrm{20m}$ 
space is established in MATLAB R2021a. In order to verify the 
effectiveness of the proposed method, the local minimum problem, target 
unreachable problem and complex obstacle environment are tested.

\begin{figure}[t]
\centering

\includegraphics[scale=0.45]{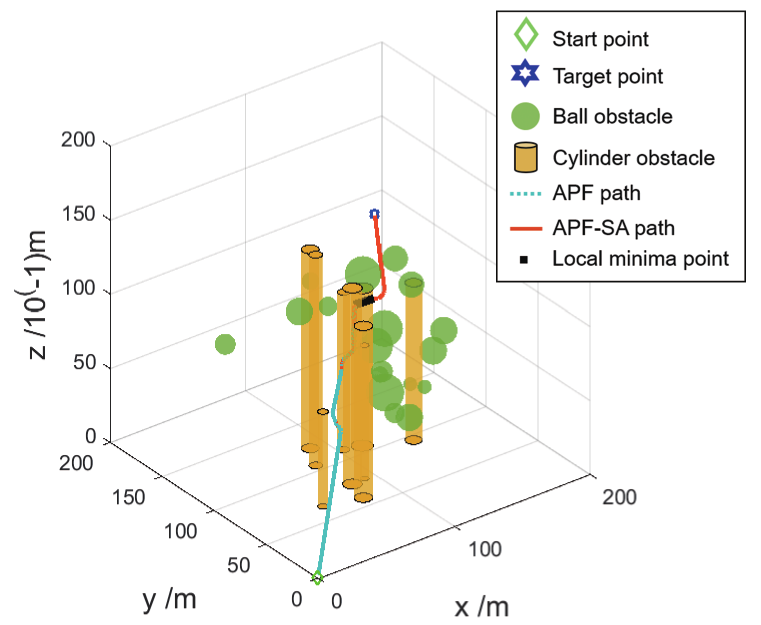}\vspace{-2mm}\textcolor{black}{\caption{\label{fig:Local minimum}Performance of local minimum with different methods.}
}

\vspace{-3mm}
\end{figure}

To evaluate the performance of the proposed improved APF-SA, 
we compare the paths generated by the traditional APF algorithm with 
those generated by APF-SA in the same fixed space. The flying 
environment in the forest 
is simulated by setting up spherical and cylindrical obstacles. As shown 
in Fig. \ref{fig:Local minimum}, oscillation points are generated in the 
path of the traditional algorithm, and it is explained that the UAV is 
trapped in a local minimum. Meanwhile, APF-SA generates random track 
points in the path, and finally successfully breaks away from the local 
minimum dilemma and reaches the target point. Thus, 
Fig. \ref{fig:Local minimum} demonstrates the efficiency and 
effectiveness of APF-SA, when facing the local minimum problem.

\begin{figure}[t]
\centering

\includegraphics[scale=0.45]{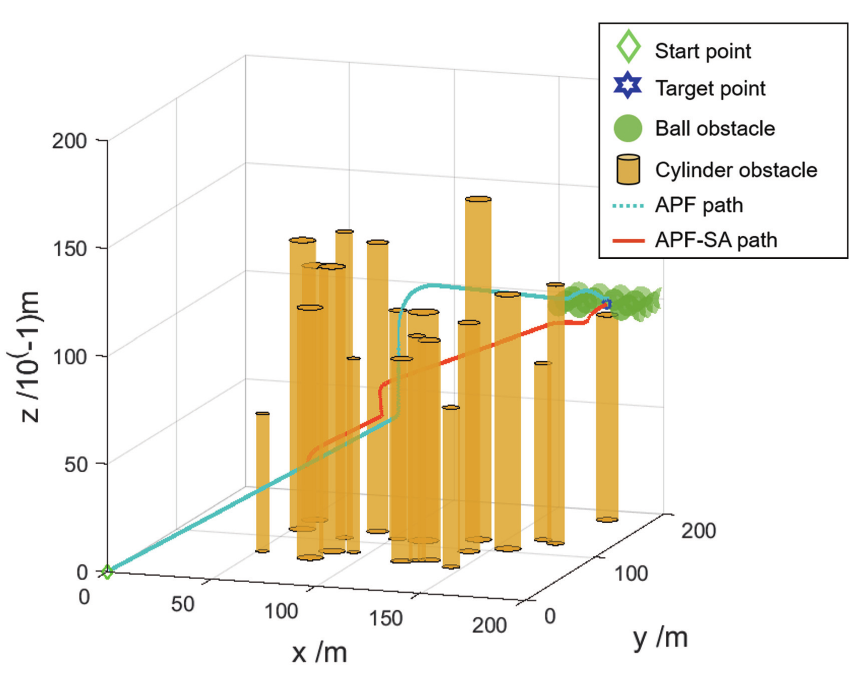}\vspace{-2mm}\textcolor{black}{\caption{\label{fig:Unreachable target}Performance of unreachable target with different methods.}
}

\vspace{-3mm}
\end{figure}

\begin{figure}[t]
    \centering
    
    \includegraphics[scale=0.45]{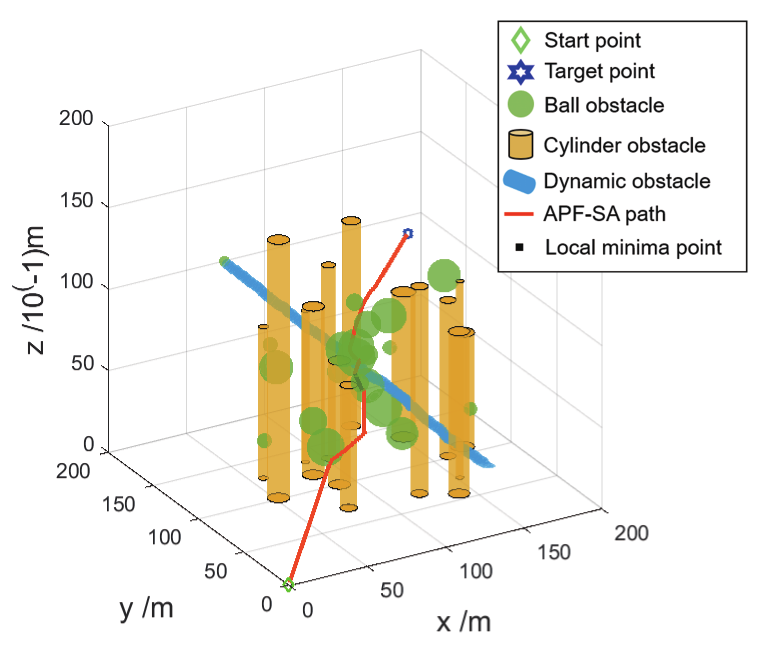}\vspace{-2mm}\textcolor{black}{\caption{\label{fig:general}Performance in complex space.}
    }
    
    \vspace{-3mm}
\end{figure}

Additionally, in Fig. \ref{fig:Unreachable target}, the path generated by 
APF-SA is more robust by modifying and optimizing the attractive 
and repulsive field functions. It is observed that in the complex
obstacle avoidance environment, the traditional algorithm can not reach 
the target point, while APF-SA can achieve obstacle avoidance and reach 
the target well. Such target reachability demonstrates the robustness of APF-SA.
Moreover, in Fig. \ref{fig:general}, when dynamic obstacles are added to the 
scene, APF-SA also has excellent obstacle avoidance 
capacity, so it has good generalization performance and 
can handle static and dynamic obstacles well.

\begin{figure}[t]
\centering

\subfloat[Path comparison map.\label{fig:comparison}]{\includegraphics[scale=0.45]{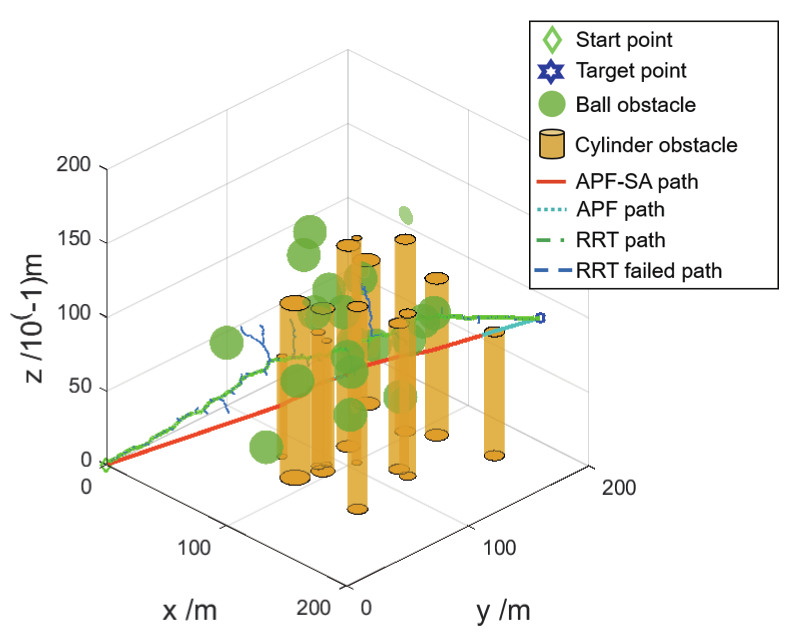}\vspace{-3mm}

}\vspace{-1mm}

\subfloat[Performance comparison.\label{fig:DA}]{\centering

\includegraphics[scale=0.47]{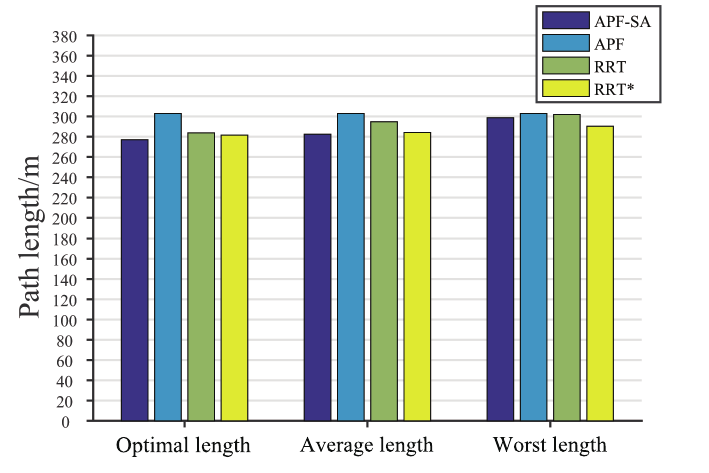}\vspace{-3mm}

}

\caption{Comparisons of multiple algorithms.\label{fig:Performance of different algorithms}}

\vspace{-3mm}
\end{figure}

As shown in Fig. \ref{fig:Performance of different algorithms}, we compare the traditional APF, RRT, RRT* and 
APF-SA algorithms. Simulation results show that the optimal path and average 
path of APF-SA are the shortest, and the worst path is slightly longer than 
that of RRT* when each algorithm runs 100 times. Therefore, APF-SA shows 
advantages in path length, time complexity, as well as resource efficiency.

\vspace{3mm}

\section{Conclusions\label{sec:Conclusions}}

In this paper, we investigated the autonomous path planning of UAV 
based on obstacle avoidance function in 3D space for 
emergency rescue in forest scenarios. We employed 
cylindrical and spherical obstacles to respectively approximate forests and no-fly 
zones in the forest, and focused on navigating through complex 
obstacle environments. To handle the issues, we designed the APF-SA to obtain the optimized path efficiently. 
Simulation results showed that compared with the traditional APF, APF-SA 
obtained a better path and solved the problem of 
unreachable target to a great extent. Simulation results showed 
that APF-SA exhibited excellent obstacle avoidance performance under both static and 
dynamic obstacle environment.\vspace{-5mm}

\textcolor{black}
{\bibliographystyle{IEEEtran}
\bibliography{ref}

\begin{thebibliography}{10}
\providecommand{\url}[1]{#1}
\csname url@samestyle\endcsname
\providecommand{\newblock}{\relax}
\providecommand{\bibinfo}[2]{#2}
\providecommand{\BIBentrySTDinterwordspacing}{\spaceskip=0pt\relax}
\providecommand{\BIBentryALTinterwordstretchfactor}{4}
\providecommand{\BIBentryALTinterwordspacing}{\spaceskip=\fontdimen2\font plus
\BIBentryALTinterwordstretchfactor\fontdimen3\font minus \fontdimen4\font\relax}
\providecommand{\BIBforeignlanguage}[2]{{%
\expandafter\ifx\csname l@#1\endcsname\relax
\typeout{** WARNING: IEEEtran.bst: No hyphenation pattern has been}%
\typeout{** loaded for the language `#1'. Using the pattern for}%
\typeout{** the default language instead.}%
\else
\language=\csname l@#1\endcsname
\fi
#2}}
\providecommand{\BIBdecl}{\relax}
\BIBdecl

\bibitem{10559211}
Z.~Lu, Z.~Jia, Q.~Wu, and Z.~Han, ``Joint trajectory planning and communication design for multiple {UAVs} in intelligent collaborative air–ground communication systems,'' \emph{IEEE Internet Things J.}, vol.~11, no.~19, pp. 31\,053--31\,067, Oct. 2024.

\bibitem{DONG2024}
C.~DONG, Y.~ZHANG, Z.~JIA, Y.~LIAO, L.~ZHANG, and Q.~WU, ``Three-dimension collision-free trajectory planning of {UAVs} based on {ADS-B} information in low-altitude urban airspace,'' \emph{Chin. J. Aeronaut.}, Apr. 2024.

\bibitem{10599389}
Z.~Jia, J.~You, C.~Dong, Q.~Wu, F.~Zhou, D.~Niyato, and Z.~Han, ``Cooperative cognitive dynamic system in {UAV} swarms: Reconfigurable mechanism and framework,'' \emph{IEEE Veh. Technol. Mag.}, vol.~19, no.~3, pp. 90--101, Sep. 2024.

\bibitem{9970002}
M.~Kim, J.~Ahn, and J.~Park, ``{TargetTree-RRT*: Continuous-Curvature Path Planning Algorithm for Autonomous Parking in Complex Environments},'' \emph{IEEE Trans. Autom. Sci. Eng.}, vol.~21, no.~1, pp. 606--617, Jan. 2024.

\bibitem{10225633}
P.~Wu, L.~Zhong, J.~Xiong, Y.~Zeng, and M.~Pei, ``{Two-level vehicle path planning model for multi-warehouse robots with conflict solution strategies and improved ACO},'' \emph{J. Intell. Connected Veh.}, vol.~6, no.~2, pp. 102--112, Jun. 2023.

\bibitem{6561444}
Y.~He, Q.~Zeng, J.~Liu, G.~Xu, and X.~Deng, ``{Path planning for indoor UAV based on Ant Colony Optimization},'' in \emph{2013 25th Chinese Control and Decision Conference (CCDC), Guiyang, China}, 2013, pp. 2919--2923.

\bibitem{10295735}
J.~Li, X.~Xiong, and Y.~Yang, ``{A Method of UAV Navigation Planning Based on ROS and Improved A-star Algorithm},'' in \emph{2023 CAA Symposium on Fault Detection, Supervision and Safety for Technical Processes (SAFEPROCESS), Yibin, China}, 2023.

\bibitem{9830995}
S.~Xie, J.~Hu, P.~Bhowmick, Z.~Ding, and F.~Arvin, ``{Distributed Motion Planning for Safe Autonomous Vehicle Overtaking via Artificial Potential Field},'' \emph{IEEE Trans. Intell. Transp. Syst.}, vol.~23, no.~11, pp. 21\,531--21\,547, Nov. 2022.

\bibitem{10250980}
Q.~Ma, M.~Li, G.~Huang, and S.~Ullah, ``{Overtaking Path Planning for CAV Based on Improved Artificial Potential Field},'' \emph{IEEE Trans. Veh. Technol.}, vol.~73, no.~2, pp. 1611--1622, Feb. 2024.

\bibitem{10033554}
X.~Chen and J.~Fan, ``{UAV trajectory planning based on APF-RRT* algorithm with goal-biased strategy},'' in \emph{2022 34th Chinese Control and Decision Conference (CCDC)}, 2022, pp. 3253--3258.

\bibitem{9538804}
Z.~Pan, C.~Zhang, Y.~Xia, H.~Xiong, and X.~Shao, ``{An Improved Artificial Potential Field Method for Path Planning and Formation Control of the Multi-UAV Systems},'' \emph{IEEE Trans. Circuits Syst. II Express Briefs Transactions on Circuits and Systems II: Express Briefs}, vol.~69, no.~3, pp. 1129--1133, Mar. 2022.

\bibitem{10404342}
J.~He, Z.~Jia, C.~Dong, J.~Liu, Q.~Wu, and J.~Liu, ``{UAV Swarm Deployment and Trajectory for 3D Area Coverage via Reinforcement Learning},'' in \emph{2023 International Conference on Wireless Communications and Signal Processing (WCSP)}, 2023, pp. 683--688.

\bibitem{9184929}
Z.~Jia, M.~Sheng, J.~Li, D.~Niyato, and Z.~Han, ``{LEO-Satellite-Assisted UAV: Joint Trajectory and Data Collection for Internet of Remote Things in 6G Aerial Access Networks},'' \emph{IEEE Internet Things J}, vol.~8, no.~12, pp. 9814--9826, Jun 2021.

\bibitem{10169744}
V.~Dubey, B.~Patel, and S.~Barde, ``{Path Optimization and Obstacle Avoidance using Gradient Method with Potential Fields for Mobile Robot},'' in \emph{2023 International Conference on Sustainable Computing and Smart Systems (ICSCSS), Coimbatore, India}, 2023, pp. 1358--1364.

\bibitem{9832753}
Q.~Fan, G.~Cui, Z.~Zhao, and J.~Shen, ``{Obstacle Avoidance for Microrobots in Simulated Vascular Environment Based on Combined Path Planning},'' \emph{IEEE Rob. Autom. Lett.}, vol.~7, no.~4, pp. 9794--9801, Oct. 2022.

\bibitem{9734752}
A.~Gottardi, S.~Tortora, E.~Tosello, and E.~Menegatti, ``{Shared Control in Robot Teleoperation With Improved Potential Fields},'' \emph{IEEE Trans. Hum.-Mach. Syst.}, vol.~52, no.~3, pp. 410--422, Jun. 2022.

\bibitem{9371396}
C.~Li, F.~You, T.~Yao, J.~Wang, W.~Shi, J.~Peng, and S.~He, ``{Simulated Annealing Particle Swarm Optimization for High-Efficiency Power Amplifier Design},'' \emph{IEEE Trans. Microwave Theory Tech.}, vol.~69, no.~5, pp. 2494--2505, May. 2021.

\bibitem{10079729}
S.~Cao, R.~Wang, and X.~Du, ``{Performance Optimization of Variable Cycle Engine Based on Improved Simulated Annealing Algorithm},'' in \emph{2022 8th International Conference on Control Science and Systems Engineering (ICCSSE)}, 2022, pp. 159--164.

\bibitem{9534874}
F.~Okoli, J.~Bert, S.~Abdelaziz, N.~Boussion, and D.~Visvikis, ``{Optimizing the Beam Selection for Noncoplanar VMAT by Using Simulated Annealing Approach},'' \emph{IEEE Trans. Radiat. Plasma Med. Sci.}, vol.~6, no.~5, pp. 609--618, May. 2022.

\end{thebibliography}
}
\end{document}